# Simple Two-wheel Self-Balancing Robot Implementation

Sayed Erfan Arefin
saarefin@ttu.edu
Texas Tech University
Lubbock, Texas, USA

## ABSTRACT

Cyber physical systems, also known as CPS, is an emerging field of technology that combines the physical and digital worlds by allowing for seamless interaction and communication between the two. One of the key characteristics of a CPS is its ability to take input from its environment and use that information to produce an output through actuators in the physical world.

A balancing robot is a prime example of a CPS, as it uses input from its sensors to continually monitor its orientation and take action to prevent falling over by generating thrust through its wheels or manipulating its inertia. In this specific project, a two-wheel self-balancing robot was developed, utilizing the concept of a reverse pendulum. A reverse pendulum by default is inherently unstable and requires an external force to maintain its balance. In this case, the balancing robot produces this external force through the use of wheels and motors.

To achieve precise balancing, stepper motors were utilized in the design of the robot. Additionally, the robot has the capability to move in four basic directions and the movement is controlled through an app connected to the robot via Bluetooth. This allows for remote control and monitoring of the robot's movements and actions. Overall, the development of this two-wheel self-balancing robot serves as a demonstration of the potential and capabilities of cyber physical systems technology.

## 1 INTRODUCTION

Cyber-physical systems (CPS) are no longer just theoretical concepts but are becoming a reality in various industries. One of the most prominent applications of CPS is in the field of robotics, where the integration of physical and digital systems allows for the creation of advanced and autonomous machines. A self-balancing robot is a prime example of a CPS, as it is a simple yet sophisticated machine that utilizes input from its environment to maintain its balance and stability [8].

There are various types of self-balancing robots, each with their own unique features and capabilities. These robots can be divided into categories based on their structure and functionality. Some examples include two-wheeled self-balancing robots, which balance on two wheels, and three-wheeled self-balancing robots, which use additional wheels for stability. Other types include self-balancing robots that use different types of sensors, such as inertial measurement units (IMUs), to monitor and control their balance and self-balancing robots that use advanced control algorithms, such as feedback control, to maintain their stability.

Regardless of their specific design, all self-balancing robots share the common goal of maintaining their balance and stability, utilizing advanced technology to achieve this goal. This makes them an excellent example of the capabilities and potential of cyber-physical systems technology.

Different types of self-balancing robots:

- Two wheel self balancing robot
- One wheel self balancing robot
- Balancing cube. etc.

A two-wheel self-balancing robot is a type of robot that utilizes two wheels that are placed vertically on the same axis. The primary goal of the robot is to maintain its balance and prevent falling over. It achieves this by constantly monitoring its orientation and making adjustments to its movement, such as moving forward or backward, in order to maintain a stable position.

Additionally, this type of robot typically employs sensors, such as inertial measurement units (IMUs), to gather data on its orientation and movement. This data is then processed by a control algorithm, which can be implemented on an onboard microcontroller such as an Arduino, to determine the necessary adjustments to be made to the robot's movement in order to maintain balance.

Furthermore, the robot can also be controlled remotely through an app connected via Bluetooth, which allows for real-time monitoring and control of the robot's movement and actions. The precise balance of the robot is achieved through the use of precise motors, such as stepper motors.

Overall, the two-wheel self-balancing robot is a prime example of the integration of advanced technology, including sensors, control algorithms, and precise motors, to achieve a specific task in this case maintaining balance [7]. It is also a perfect example of cyber physical systems, which seamlessly integrates the physical and digital worlds.A one wheel self balancing robot balances it self in a single wheel. It can do that by moving forward and backward and also by maintaining its inertia. The inertia can be maintained by spinning a heavy disc very fast. The concept of preserving inertia by spinning disk is also used in a balancing cube. A balancing cube actually balances it self by one of the corners. There are 3 disks placed in the 3 axis of the robot. By spinning the disks in different rotational speed, the inertia is preserved by the robot.

In this project the two wheel self balancing robot was developed.

The paper is organized as the following sections. First the model is discussed. Secondly, the methods used to develop this robot is discussed. Afterwards, the results were discussed. Following this the discussion and conclusion can be found.

## 2 MODEL

At first, UML diagrams were created to model the robot. The use case diagram of the robot is shown in Figure 1. The class diagram is given in Figure 3. The state chart diagram is shown in Figure 4. The circuit wiring of the robot is shown in Figure. 2.

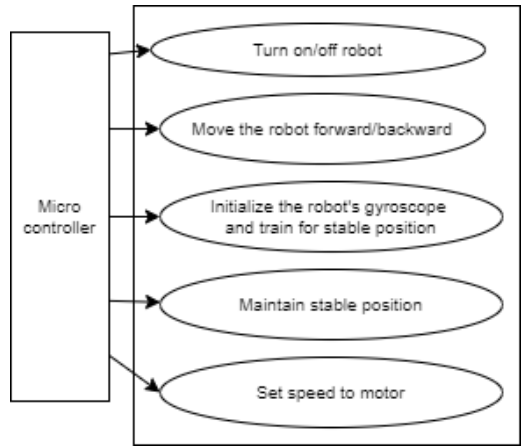

**Figure 1: Use case diagram of the self balancing robot**

The following parts were required for this project.

- Two wheels
- Stepper motor: Nema 17 [5]
- Stepper motor driver: A4988 [6]
- Bluetooth module: HC-06 [3]
- Current limiter: LM2596 [4]
- Voltage regulator: 7905 [1]
- Stepper motor mount
- Battery: 5100 mAh Lithium Polymer (LiPo) Battery
- Arduino UNO

## 3 METHOD
### 3.1 Structural design

The structure design of the robot is based on the concept of a reverse pendulum. In a traditional pendulum, the pivot point is located above the center of mass, but in the case of a reverse pendulum, the pivot point is located below the center of mass. This design choice is intentional, as it makes the structure inherently unstable and will cause the robot to fall over unless external forces are used to recover from any tilting motion.

The motion of a reverse pendulum is illustrated in Figure 5. The figure shows how the robot's center of mass is located above the pivot point, and how the robot needs to use external forces to maintain its balance and prevent falling over.

The use of a reverse pendulum design is a key aspect of the robot's functionality, as it allows the robot to maintain its balance and recover from any tilting motion. The robot's control algorithm uses sensor data to monitor its orientation and apply the necessary external forces to keep the robot upright.

Overall, the structure design of the robot plays a critical role in its operation, as it determines the robot's stability and the forces required to keep it balanced. The reverse pendulum design is a unique feature of the robot which helps it to function effectively and overcome the instability caused by having the pivot point below the center of mass.

### 3.2 Control loop

In this project, the PID or proportional–integral–derivative controller was used as the control loop.

The ideal state of the robot is standing up. This means the robot will not be tilting. The MPU 6050 reads the position of the robot and calculates the error e(t). The desired state is the position you want your robot to be in. In the case of the balancing robot, that would be straight up and down. There are 3 constants defined in the controller. Those are defined as follows, Kp, Ki, and Kd. The proportional part is as follows, *Kp * e(t)*. The integral part is: *Ki * integral e(t)* and the derivative part is: *Kd * d/dt e(t)*. These terms are added and fed to the system. The final signal is as follows.

$$signal = K_p . e(t) + K_i \int e(t) \frac{d}{dt} + K_d \frac{d}{dt} e(t)$$

The control signal tells the Arduino how fast to move the motors in order to balance the robot. Then, the MPU 6050 sensor reads the robot's position. Afterward, the new current position is sent to the Arduino. Then, the Arduino calculates the error and does all these steps again. This is also known as feedback. This runs in the loop which makes this control loop a closed-loop system.

The proportional term makes a change to the output signal that is proportional to the current error value. Meaning, the proportional gain will make a larger change in the signal if the error is large and if the error is small it will do otherwise. Larger values usually mean quicker response since the larger the error, the larger the Proportional term correction signal. But if the proportional gain is very large, it will make the robot oscillate which will lead to instability. If the value is too small, the robot will drift away and fall. The further away from the desired position the robot is, the larger the proportional term required to correct the position.

The Integral term is usually proportional to the duration of the existence of the error in the system. The integral term can accelerate the movement of the process toward the desired position. This also helps terminate the residual steady-state errors. The steady-state errors occur with a proportional-only controller. The integral term takes the errors and adds those up over time until it is large enough to make a difference. Then, it corrects the error. For example, given a position, the MPU should read 171 degrees to be balanced, but reads 171.1. This error of 0.1 might not be enough to move the motors. But over time this error gets accumulated. if the MPU reads that value 100 times in a second, the error of .1 degree becomes 10 degrees of cumulative error over time. This is significant enough to move the motors. A large Ki will help the robot to steady itself very fast. This can also help terminate drift.

Finally, the Kd term is also known as the derivative term. The controller output is proportional to the rate of change of the measurement or error. The derivative term slows the rate of change of the controller output as it approaches the desired position. This helps prevent the robot to overshoot itself from the desired set point [9].

The PID can be illustrated in Figure 6. In the Arduino code, the orca timer was used in CTC mode. This counter was used to identify

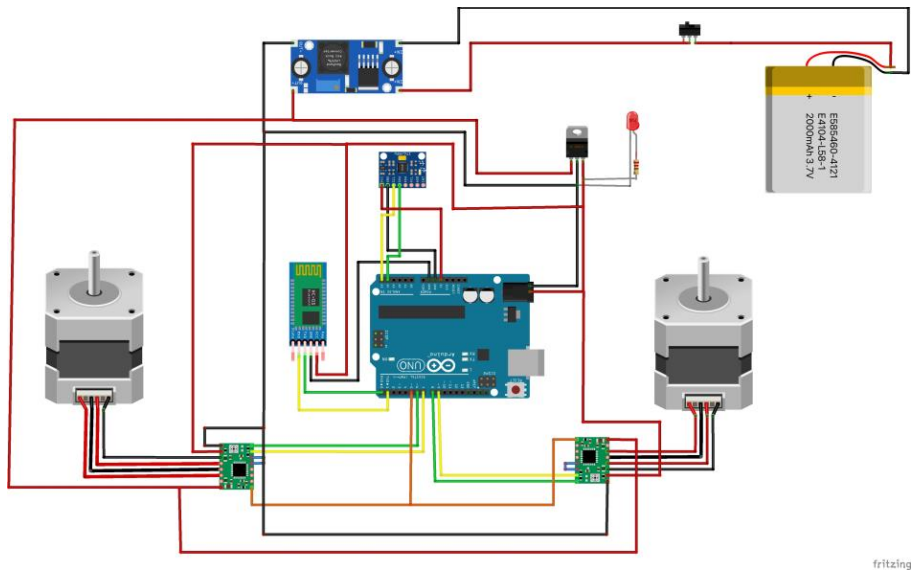

**Figure 2: Wiring of the self balancing robot. This image was created using the fritzing tool [2]**

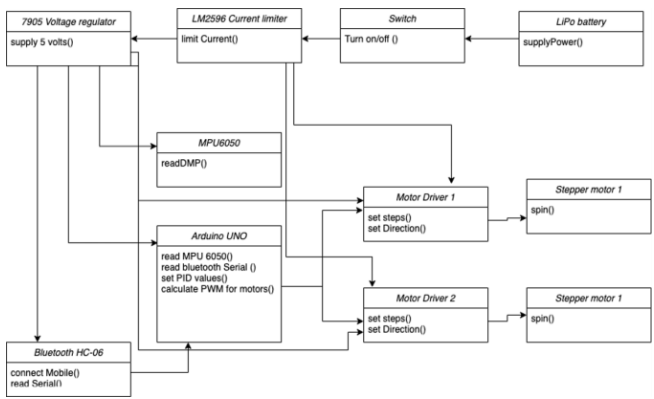

**Figure 3: Class diagram of the self-balancing robot**

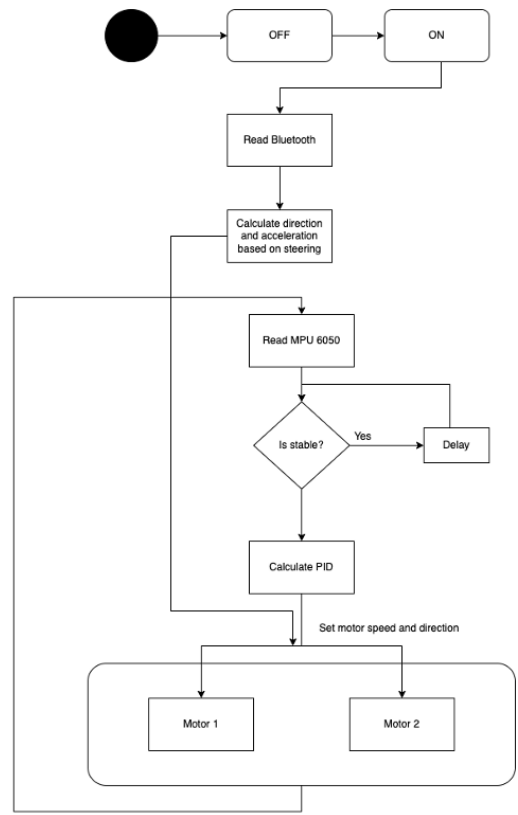

**Figure 4: State chart diagram of the self balancing robot**

time differences. Based on this, the integration and differentiation were calculated.

## 3.3 Power distribution

The power distribution of a robot is a critical aspect of its design and operation. Each component of the robot has a specific power rating, and if the power supplied to these components is outside of this range, it can lead to damage or malfunction. In this case, the power source of the robot is a 5100 mAh Lithium Polymer (LiPo) battery.

To ensure proper and stable power distribution, the system includes two components that help stabilize the power. The first is a current limiter, specifically an LM 2596, which steps down the current to 3 amps. This stepped-down current is then fed to the stepper motor power ports of the motor driver.

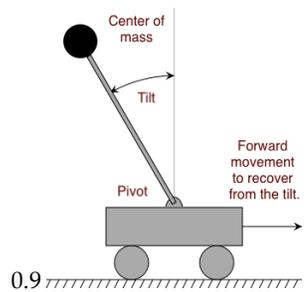

Figure 5: Reverse Pendulum

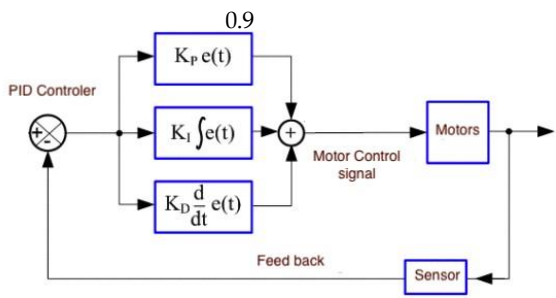

Figure 6: PID control loop

In parallel to this, there are three voltage regulators placed that use a 7905 MOSFET to limit the voltage to 5v. These voltage regulators also have a current rating of 800mA. This 5-volt power is then fed to the 5-volt components of the robot, such as the Arduino, Bluetooth module (HC-06), MPU 6050, and motor driver power. This ensures that each component is receiving the appropriate amount of power to function properly and avoid damage.

Overall, the power distribution system plays a crucial role in the robot's design and operation by ensuring the stable and appropriate power supply to the different components of the robot.

### 3.4 Movement control

The movement of the self-balancing robot is achieved by introducing small adjustments to the motors. The robot's desired location, known as the set point, is changed to achieve movement. For example, if the robot needs to move forward, the set point is shifted slightly forward, tricking the PID loop to rebalance itself and move forward. Similarly, for a backward movement, the set point is shifted backward.

To achieve smooth movement, another PID loop is implemented in the system that handles the velocity of the robot. This helps to stabilize the robot even more when movement is introduced on top of its primary task of self-balancing. The robot can also move to the right by rotating the left wheel and move to the left by rotating the right wheel.

The Bluetooth module is connected to the Arduino and reads the serial input, if available, on every loop. This reads for new movement instructions and sets the desired position based on that. The desired position can also be set through an android app that connects to the robot via Bluetooth. The app has a joystick interface that calculates the velocity and steering of the robot and sends it over Bluetooth to the robot. On every loop, this input is read and parsed to the correct variables for the movement of the robot. Overall, the system uses advanced control algorithms and sensor input to achieve precise movement and stabilization of the robot.

## 4 RESULTS

In this section, the results of the project is discussed. At this stage, the robot was fully constructed. The components powered up and performed properly when tested individually. Afterwards, the robot was tuned. The final PID values were as follows: Kp = 1150.0, Kd = 157.5 and Ki = 0.12. The robot was balancing and was stable at this time. If an external force on induced in the robot, it soon re overed from the tilt and kept balancing. When large amount of force is applied on the robot, it tried to balance but eventually fell off. The main reason of this is the small form factor of the robot. The blue tooth connected properly and by moving the joystick in the mobile app, the robot moved in four different directions.

## 5 DISCUSSION

In this section, I will delve into the technical difficulties encountered during the robot's development process.

### 5.1 PID tuning

THE PID tuning of the robot is very important in this project. Very small error in the tuning can lead to sever consequences. This tuning was done in a trial and error manner.

At first the P and D were tuned and after wards the I was tuned. The values for the P, I and D will be different for every robot, as the center of gravity will be different for every robot. The P value is considered as the current value. At first the P value is set and the robot actions are observed. Based on this, the P values are changed over time to find a suitable value of P. If the P value is too small the robot will never cross the center of gravity. Meaning if the robot is tilted it will try to recover the tilt but it will never settle and will keep on drifting. At some point it may fall over. If the P value is too high, the robot will try to recover a tilt but it will always over shoot it. As a result the robot will have an oscillating motion.
Once the robot has a tolerable oscillating motion, the D value is set. The D value is considered as the derivative. As the P value forces the robot to recover from a tilt, the D value helps reduce the acceleration as the robot approaches the center of gravity. A proper D value will help the robot to balance at the right time while recovering from a tilt. Too small value will not stop the robot while recovering from a tilt and the robot will keep on oscillating. If the D value is too large, the robot will stop before recovering from a tilt, causing the robot to fall over.
The I or the integral value helps the robot the correct its position in the precision context. After setting the P and D values, the I value is set to give the robot a precise position.

## 5.2 Structure related discussion

The robot is constructed with an aluminum chassis, which provides a sturdy and durable structure. While it could be developed using plastic as well, the focus of the design is to have a robust and reliable structure. Several screws were used to mount the various components of the robot, ensuring a secure and stable attachment.

This is especially important during the tuning process, as the robot may make unusual movements. A poorly constructed structure can lead to structural failures and compromise the performance of the robot. To enhance the stability and balance of the robot, the battery is placed at the top of the robot, which helps to keep the center of gravity at the top. This improves the stability of the robot, particularly when it's in motion.

The MPU (Motion Processing Unit) is placed at the bottom of the robot to capture most of the motion produced by the motors as it attempts to recover from any tilt or imbalance. This positioning of the MPU allows for accurate and precise monitoring of the robot's movements and facilitates the control algorithms in maintaining balance.

Overall, the design of the robot's structure and placement of its components is critical in ensuring its stability and robustness, which is essential for its proper functioning and performance.

## 5.3 Testing and calibrating individual components

After completing the construction and wiring of the robot, it is crucial to thoroughly test each component individually, even if they were functioning properly before the construction process. This ensures that all parts are working correctly and that any issues can be identified and addressed before the robot is put into operation. One important aspect of the testing process is calibrating the MPU (Motion Processing Unit). The MPU uses sensors to gather data on the robot's orientation and movement, and it is important to ensure that these sensors are properly calibrated. The offsets of the sensor may change after installation, and if they are not calibrated correctly, it can lead to inaccurate readings and affect the robot's performance.

The calibration process involves adjusting the sensor's settings to match its specific environment and conditions, and it should be done carefully and precisely to ensure accurate readings. This process may have to be repeated multiple times to achieve the desired level of accuracy.

Overall, the testing and calibration of the robot's components are vital steps in the development process, as it ensures that the robot will function properly and perform to its full potential.

*5.3.1 MPU Calibration.* To accurately measure the orientation of the robot using the MPU 6050 sensor, it is crucial to calibrate the sensor. One of the key reasons for this is that every robot has a unique structure, even if it is an identical copy of another. Therefore, the position of the sensor may vary between robots, and thus, the sensor must be calibrated to account for these differences.

The MPU6060 library comes with a calibration option, which allows for the calibration of the sensor. During the calibration process, the robot is placed in a vertical position, and an object is placed beneath the robot to ensure that it is completely vertical. After the calibration, the sensor provides 6 offset data for 2 different contexts, including the acceleration and gyroscope, which both provide 3-axis information for the offset.

These offset data are then used in the main code to properly balance the robot. This ensures that the robot's orientation is accurately measured and controlled, which is crucial for its stability and proper functioning. Overall, the calibration of the sensor is a vital step in the development process, as it helps to achieve precise and accurate measurements of the robot's orientation.

*5.3.2 MPU testing.* The MPU 6050 sensor is equipped with a DMP (Digital Motion Processor) which is used to merge the values of acceleration and gyroscope and send the data to the arduino via I2C communication. In this project, the wire library was utilized to establish communication between the DMP of the MPU and the arduino.

It is important to test the DMP independently, as changes in the power distribution of the robot after its construction may affect the MPU. By testing the DMP separately, it is possible to verify that it initializes properly and provides accurate output over the serial. This step is important to ensure that the DMP is functioning properly, and that the robot will be able to properly process the sensor data to maintain its balance and stability.

Overall, the DMP is a crucial component of the robot, and its proper functioning is essential for the robot's ability to accurately measure and control its orientation. The testing of the DMP independently helps to ensure that it is working as intended and that the robot will be able to perform to its full potential.

*5.3.3 Motors testing.* The motors used in this robot are bipolar stepper motors, which are connected to the arduino through a motor driver. To ensure the proper functioning of the robot, it is important to test the motors individually and together in terms of direction and stepping.

After the final wiring and physical setup of the robot, the motors were tested to ensure they were functioning properly. This includes testing their ability to rotate in the correct direction, as well as their ability to take precise steps, which is crucial for the robot's ability to maintain its balance and move smoothly.

The testing of the motors is an essential step in the development process, as it helps to ensure that the robot will be able to perform its intended tasks. Any issues with the motors can lead to poor performance, and potentially even damage to the robot. By testing the motors individually and together, it is possible to identify and address any issues before the robot is put into operation.

Overall, the motors play a crucial role in the robot's operation, and proper testing is important to ensure they are functioning correctly and to the fullest potential.

*5.3.4 Bluetooth Testing.* The Bluetooth module is connected to the arduino through a serial connection. In order to establish communication between the two devices, it is important to properly configure the Bluetooth module. One of the key configurations is setting the baud rate to 115200.

This is accomplished using AT commands, which are a set of instructions used to program and configure the Bluetooth module. These commands allow for various settings to be adjusted, including the baud rate, to ensure that the module is properly configured for the intended use case. In this case, the baud rate is set to 115200, which ensures that the data is transmitted at a high enough speed to maintain smooth communication between the arduino and the Bluetooth module.

The use of AT commands is an essential step in the setup process, as it allows for precise configuration of the Bluetooth module and ensures that it will function as intended. The correct configuration of the baud rate is important for the smooth communication between the arduino and the bluetooth module.

Overall, the Bluetooth module plays an important role in the robot's operation, as it allows for remote control and monitoring of the robot. Proper configuration using AT commands is crucial to ensure that the module functions correctly and enables smooth communication between the robot and the controlling device.

## 6 CONCLUSION

A self-balancing robot is a prime example of a cyber-physical system (CPS) due to its ability to seamlessly integrate the physical and digital worlds. In this project, the concept of a reverse pendulum was utilized to develop a self-balancing robot, covering all aspects of the design process from modeling to control. The robot is designed to balance itself on two wheels and has the capability to recover from any external forces that may cause it to lose balance.

Control information is sent to the robot through an android app via Bluetooth, which allows for remote control of the robot's movements. The robot can move in four basic directions: forward, backward, left, and right. The code used in the arduino, which forms the brain of the robot, is provided in the appendix A section of the project.

The project also includes a discussion on the crucial aspects of the robot such as tuning, wiring, structural design, and testing. All of these are important factors in the development of a self-balancing robot, and they have been discussed in great detail before the conclusion of the project.

## A CODE

The code used in the arduino uno are given below.

*A.0.1 main.ino.* include <Wire.h>
include <Arduino.h>
include "I2Cdev.h"
include "MPU6050_6Axis_MotionApps20.h"

include "PID.h"

// PORTB, bit 1, PB1
define MOT_A_STEP 9
define MOT_A_DIR 8

// PORTD, bit 7, PD7
define MOT_B_STEP 7
define MOT_B_DIR 6

define INTERRUPT_PIN 2

define PPR 1600

define TICKS_PER_SECOND 50000 // 50kHz
define PULSE_WIDTH 1

//define MAX_ACCEL (200)
//define ANGLE_Kp 450.0
//define ANGLE_Kd 30.0
//define ANGLE_Ki 0.0
// //define VELOCITY_Kp 0.007
//define VELOCITY_Kd 0.0
//define VELOCITY_Ki 0.0005

//define MAX_ACCEL (200)
//define ANGLE_Kp 740.0
//define ANGLE_Kd 145.0
//define ANGLE_Ki 190.0
// //define VELOCITY_Kp 0.005
//define VELOCITY_Kd 0.0
//define VELOCITY_Ki 0.0005

//define MAX_ACCEL (200)
//define ANGLE_Kp 1040.0
//define ANGLE_Kd 185.0
//define ANGLE_Ki 290.0
// //define VELOCITY_Kp 0.005
//define VELOCITY_Kd 0.0
//define VELOCITY_Ki 0.0005

```
//// best one
define MAX_ACCEL (200)
define ANGLE_Kp 1150.0
define ANGLE_Kd 157.5
define ANGLE_Ki 0.12

define VELOCITY_Kp 0.1
define VELOCITY_Kd 0.002
define VELOCITY_Ki 0.0005

//define MAX_ACCEL (200)
//define ANGLE_Kp 500.0
//define ANGLE_Kd 10
//define ANGLE_Ki 0.01
// //define VELOCITY_Kp 0.0001
//define VELOCITY_Kd 0.00
//define VELOCITY_Ki 0.00005

define WARMUP_DELAY_US (5000000UL)

define ANGLE_SET_POINT (2.0 * DEG_TO_RAD)

define OUTPUT_READABLE_YAWPITCHROLL
// define COUNT_LOOP
define LOGGING_ENABLED false

//define NANO_BLE

/* BLE communication-related params */v define MAX_PACKET_SIZE 96
define DIVISOR 10000.0

char packet[MAX_PACKET_SIZE];
uint8_t packet_size = 0;

MPU6050 mpu;
bool dmpReady = false;
uint8_t mpuIntStatus;
uint8_t fifoBuffer[64];

Quaternion q; // [w, x, y, z] quaternion container
VectorFloat gravity; // [x, y, z] gravity vector
float ypr[3]; // [yaw, pitch, roll] yaw/pitch/roll
// container and gravity vector

volatile bool mpuInterrupt = false; // indicates whether MPU interrupt pin has gone high
void dmpDataReady()
mpuInterrupt = true;

PID anglePID(ANGLE_Kp, ANGLE_Kd, ANGLE_Ki, ANGLE_SET_POINT);
PID velocityPID(VELOCITY_Kp, VELOCITY_Kd, VELOCITY_Ki, 0.0);

float pid_settings[6] =
ANGLE_Kp, ANGLE_Kd, ANGLE_Ki,
VELOCITY_Kp, VELOCITY_Kd, VELOCITY_Ki
;

float ref_velocity = 0.0f;
float ref_steering = 0.0f;
float steering = 0.0f;

String theString;

volatile unsigned long currentTickLeft = 0UL;
volatile unsigned long currentTickRight = 0UL;
volatile unsigned long ticksPerPulseLeft = UINT64_MAX;
volatile unsigned long ticksPerPulseRight = UINT64_MAX;
volatile float accel = 0.0;
volatile float velocity = 0.0;

bool isBalancing = false;

float angle = 0.0;
float targetAngle = ANGLE_SET_POINT;

float targetVelocity = 0.0;

unsigned long lastUpdateMicros = 0;

void send_float_array(float *a, uint8_t size);
void parse_float_array(char *p, uint8_t p_size, float *dest);
void parse_settings(char *p, uint8_t p_size);
void parse_control(char *p, uint8_t p_size);
void handle_packet(char *p, uint8_t p_size);
void splitSerialInputPID();

void splitSerialInputMove();

String createSerailOut();

void splitSerialInputMove()
int commaIndex = theString.indexOf(',');
int secondCommaIndex = theString.indexOf(',', commaIndex + 1);

ref_velocity = theString.substring(0, commaIndex).toFloat();
ref_steering = theString.substring(commaIndex + 1, secondCommaIndex).toFloat();

String createSerailOut()

return "x" + String(ref_velocity)+"," + String(ref_steering) ;

void initMPU()
const int16_t accel_offset[3] = -5508, -730, 666 ;
const int16_t gyro_offset[3] = 7, 18, -10 ;

Wire.begin();

Wire.setClock(400000);
mpu.initialize();
```

```
   pinMode(INTERRUPT_PIN, INPUT);

   if (!mpu.testConnection())
      Serial.println("MPU6050 connection failed");
      while(1)
   else
      Serial.println("MPU6050 connection done");

   mpu.dmpInitialize();

   mpu.setXGyroOffset(gyro_offset[0]);
   mpu.setYGyroOffset(gyro_offset[1]);
   mpu.setZGyroOffset(gyro_offset[2]);
   mpu.setXAccelOffset(accel_offset[0]);
   mpu.setYAccelOffset(accel_offset[1]);
   mpu.setZAccelOffset(accel_offset[2]);

   mpu.setDMPEnabled(true);
   attachInterrupt(digitalPinToInterrupt(INTERRUPT_PIN), dmpDataReady,
RISING);

void initMotors()
   pinMode(MOT_A_DIR, OUTPUT);
   pinMode(MOT_A_STEP, OUTPUT);
   pinMode(MOT_B_DIR, OUTPUT);
   pinMode(MOT_B_STEP, OUTPUT);
   digitalWrite(MOT_A_STEP, LOW);
   digitalWrite(MOT_B_STEP, LOW);

   void setTimer1(int ocra)
   TCCR1A = 0; // set entire TCCR1A register to 0
   TCCR1B = 0; // same for TCCR1B
   TCNT1  = 0; // initialize counter value to 0

   OCR1A = ocra;
   TCCR1B |= (1 « WGM12); // turn on CTC mode
   TCCR1B |= (1 « CS11); // set prescaler to 8
   TIMSK1 |= (1 « OCIE1A); // enable timer compare interrupt

   void setTimers()
   cli();
   // setTimer1(49); // 40kHz
   setTimer1(39); // 50kHz
   sei();

   bool mpuUpdate()
   if (mpuInterrupt  mpu.dmpGetCurrentFIFOPacket(fifoBuffer))
      mpu.dmpGetQuaternion(q, fifoBuffer);
      mpu.dmpGetGravity(gravity, q);
      mpu.dmpGetYawPitchRoll(ypr, q, gravity);
      mpuInterrupt = false;
      return true;

   return false;

   unsigned long getTicksPerPulse(float velocity)
   if (abs(velocity) < 1e-3)
      // TODO: disable motor
      return UINT64_MAX;
   else
      return (uint64_t)(2.0 * PI * TICKS_PER_SECOND / (abs(velocity) *
PPR)) - PULSE_WIDTH;

   void updateVelocity(unsigned long nowMicros)
   // static unsigned long counter = 0;
   // static unsigned long sum = 0;

   static unsigned long timestamp = micros();
   if (nowMicros - timestamp < 100 /* 10kHz */)
      return;

   // sum += (nowMicros - timestamp);
   // counter++;
   // if (counter >= 1000)
   // // //Serial.println(((float)(sum)) / counter);
   // counter = 0;
   // sum = 0;
   //

   float dt = ((float) (nowMicros - timestamp)) * 1e-6;
   velocity += accel * dt;

   float leftVelocity = velocity - steering;
   float rightVelocity = velocity + steering;
   ticksPerPulseLeft = getTicksPerPulse(leftVelocity);
   ticksPerPulseRight = getTicksPerPulse(rightVelocity);

   if (leftVelocity > 0)
      digitalWrite(MOT_A_DIR, HIGH);
   else
      digitalWrite(MOT_A_DIR, LOW);

   if (rightVelocity > 0)
      digitalWrite(MOT_B_DIR, HIGH);
   else
      digitalWrite(MOT_B_DIR, LOW);

   timestamp = nowMicros;

   void updateControl(unsigned long nowMicros)
   /* Wait until IMU filter will settle */
   if (nowMicros < WARMUP_DELAY_US)
      return;
```

```
    static unsigned long timestamp = micros();
if (nowMicros - timestamp < 1000 /* 1kHz */)
return;

if (!mpuUpdate())
return;

angle = ypr[1];

    float dt = ((float) (nowMicros - timestamp)) * 1e-6;

    if (abs(angle - targetAngle) < PI / 18)
// //Serial.println("isBalancing=true");
isBalancing = true;

    if (abs(angle - targetAngle) > PI / 4)
// //Serial.println("isBalancing=false");
isBalancing = false;
accel = 0.0;
velocity = 0.0;

    if (!isBalancing)
return;

targetAngle = -velocityPID.getControl(velocity, dt);
anglePID.setTarget(targetAngle);

    accel = anglePID.getControl(angle, dt);
accel = constrain(accel, -MAX_ACCEL, MAX_ACCEL);

    timestamp = nowMicros;

    void setup()
Serial.begin(115200);
while(!(Serial.available()))

    // pinMode(13, OUTPUT);
initMPU();
setTimers();
initMotors();

    /** * Stepper control interrupt handler */ ISR(TIMER1_COMPA_vect)

Serial.println("Interrupt stepper");
if (currentTickLeft >= ticksPerPulseLeft)
currentTickLeft = 0;

    if (currentTickLeft == 0)
PORTD |= _BV(PD7); // digitalWrite(MOT_B_STEP, HIGH);
else if (currentTickLeft == PULSE_WIDTH)
PORTD = _BV(PD7); // digitalWrite(MOT_B_STEP, LOW);

    currentTickLeft++;

    if (currentTickRight >= ticksPerPulseRight)
currentTickRight = 0;

    if (currentTickRight == 0)
PORTB |= _BV(PB1); // digitalWrite(MOT_A_STEP, HIGH);
else if (currentTickRight == PULSE_WIDTH)
PORTB = _BV(PB1); // digitalWrite(MOT_A_STEP, LOW);

    currentTickRight++;

    void loop()
unsigned long now = micros();
updateVelocity(now);
updateControl(now);

    ifdef COUNT_LOOP
static unsigned long last_ts = micros();
static unsigned long counter = 0;

    counter++;
if (now - last_ts >= 1000000)
// //Serial.print ("counter: ");
// //Serial.println(counter);
counter = 0;
last_ts = now;

endif

    if (Serial.available())
theString = Serial.readString();

    if (!theString.equals("xxxxx"))
if (theString.substring(0, 3).equals("mov"))
theString = theString.substring(0, 3);
splitSerialInputMove();
createSerailOut();

    static const float a = 0.99;
targetVelocity = a * targetVelocity + (1.0 - a) * ref_velocity;
velocityPID.setTarget(targetVelocity);

    steering = a * steering + (1.0 - a) * ref_steering;
```

## A.1  PID.h

```
ifndef PID_H
define PID_H
```

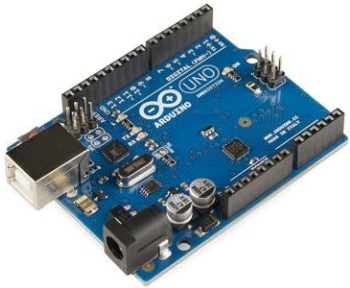

**(a) Arduino**

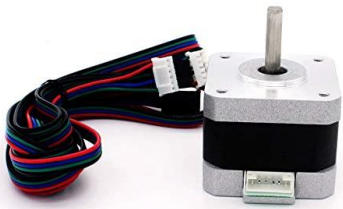

**(b) Stepper motor Nema 17**

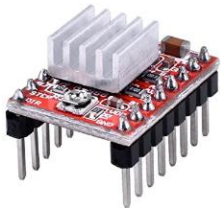

**(c) A4988 Stepper motor**

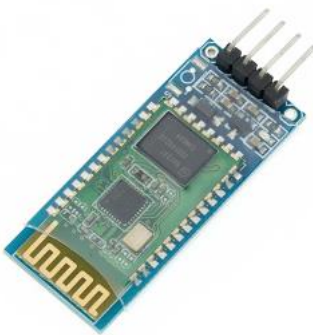

**(d) Bluetooth HC-06**

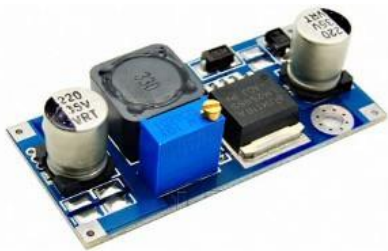

```
class PID
public:
    PID(float Kp, float Kd, float Ki, float target);
    float getControl(float value, float dt_seconds);
    void setSettings(float Kp, float Kd, float Ki);
    void setTarget(float target);
private:
    float _Kp;
    float _Kd;
    float _Ki;
    float _lastError;
    float _integralError;
    float _target;
    float _last_value;
    bool _has_last_value = false;
;

endif
```

## A.2 PID.cpp

```
include "PID.h"

PID::PID(float Kp, float Kd, float Ki, float target)
    _Kp = Kp;
    _Kd = Kd;
    _Ki = Ki;
    _target = target;

float PID::getControl(float value, float dt_seconds)
    float lastValue = _has_last_value ? _last_value : value;
    float error = _target - value;
    float de = -(value - lastValue) / dt_seconds;
    _integralError += _Ki * error * dt_seconds;
    _lastError = error;
    _last_value = value;
    _has_last_value = true;
    return (_Kp * error + _Kd * de + _integralError);

void PID::setSettings(float Kp, float Kd, float Ki)
    _Kp = Kp;
    _Kd = Kd;
    _Ki = Ki;

void PID::setTarget(float target)
    _target = target;
    _integralError = .0;
```